\documentclass[conference,10pt,letterpaper]{IEEEtran}
\usepackage{amsmath,amssymb,amsfonts}
\usepackage{graphicx}
\usepackage{booktabs}
\usepackage{array}
\usepackage{url}
\usepackage{cite}
\usepackage{balance}

\DeclareFontShape{OT1}{ptm}{m}{scit}{<->ssub*ptm/m/sc}{}

\newcommand{\FBS}{\textsc{FBS}}
\newcommand{\simdex}{\textsc{SimDex}}
\newcommand{\simtune}{\textsc{SimTune}}
\newcommand{\deltaiot}{\textsc{DeltaIoT}}
\newcommand{\mrubis}{m\textsc{RUBiS}}
\newcommand{\switchenv}{\textsc{Switch}}
\newcommand{\eopen}{\texttt{e\_open}}
\newcommand{\efbs}{\texttt{fbs\_last\_safe}}
\newcommand{\edeadline}{\texttt{e\_deadline}}
\newcommand{\emarg}{\texttt{margin\_only}}
\newcommand{\evol}{\texttt{vol\_only}}
\newcommand{\refopen}{\emph{reference-open} (\eopen{})}
\newcolumntype{P}[1]{>{\raggedright\arraybackslash}p{#1}}

\newif\ifshowfallback
\showfallbackfalse
\newcommand{\fbopen}{\textbf{TBD}}
\newcommand{\fbfbs}{\textbf{TBD}}
\newcommand{\fbmargin}{\textbf{TBD}}
\newcommand{\fbvol}{\textbf{TBD}}
\newcommand{\fbdeadline}{\textbf{TBD}}

\title{Approved Too Late: Verdict Staleness\\
in LLM-Guarded Self-Adaptive Systems}

\author{%
\IEEEauthorblockN{Ilai Shraga}
\IEEEauthorblockA{University of Cambridge\\
Cambridge, United Kingdom\\
\texttt{is628@cam.ac.uk}}
\and
\IEEEauthorblockN{Roei Eshel}
\IEEEauthorblockA{Maccabim-Re'ut High School\\
Modi'in-Maccabim-Re'ut, Israel\\
\texttt{roeieshel@gmail.com}}
\and
\IEEEauthorblockN{Lior Gorelik}
\IEEEauthorblockA{The Open University of Israel\\
Raanana, Israel\\
\texttt{golior32@365.openu.ac.il}}
}

\begin{document}
\maketitle

\begin{abstract}
A large language model (LLM) guardrail for a self-adaptive system (SAS) may issue an approval that is correct at check time but stale by actuation. This creates an Execute-stage time-of-check to time-of-use (TOCTOU) hazard. We study verdict freshness: whether a guardrail verdict remains valid when used. We distinguish three quantities that answer different questions: all-candidate verdict change under fixed-action replay, oracle-labeled approval expiry on recorded closed-loop trajectories, and judge-conditioned use-time invalidity. Across five reproducible SAS environments, all-candidate verdict-change rates span $5.3$--$48.4\%$ at a common replay shift of eight simulator steps. We introduce the Freshness-Bounded Shield (\FBS{}), which estimates each approval's validity horizon from its safe-side margin and recent feature volatility, without an explicit plant-dynamics model. Using fixed settings documented in the artifact, \FBS{} reduces oracle-labeled approval-expiry rates from $3.4$--$24.7\%$ to $0$--$1.8\%$ at the same shift. A separate audit of four LLM judges finds nonzero judge-conditioned use-time invalidity in every approval stream. We formulate a freshness contract: every approval must be correct at check time and remain valid at use time.
\end{abstract}


\section{Introduction}\label{sec:intro}

An Execute-stage guardrail can return a correct verdict on the context--action pair it checks and still authorize an action that is inadmissible when applied. In a self-adaptive system (SAS), a controller observes the plant, proposes an action, and relies on a large language model (LLM) guardrail to approve or reject it before actuation. The plant continues to evolve while the verdict is computed and delivered. By actuation, the checked context may no longer describe the current plant state, and the action may no longer be admissible. This creates an assurance gap between check-time correctness and use-time validity.

\begin{figure*}[t]
  \centering
  \includegraphics[width=0.84\textwidth]{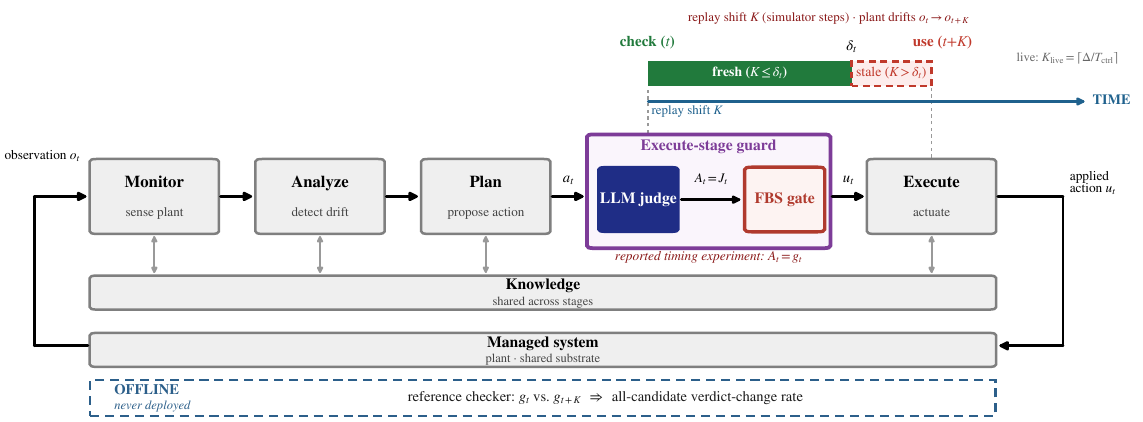}
  \caption{Pipeline view of Execute-stage verdict freshness in the MAPE-K loop. The upper path depicts the intended Judge$+$\FBS{} deployment: the LLM judge evaluates check-time context $x_t=(o_t,h_t)$ and candidate action $a_t$, and supplies upstream approval $A_t=J_t$. The reported oracle-labeled timing experiment instead sets $A_t=g_t$ to isolate temporal expiry from check-time judge error. During the check--use window, the plant evolves to the later recorded context $x_{t+K}$. For an upstream-approved candidate, \FBS{} applies $a_t$ as $u_t$ only if its safe-side margin is positive and replay shift $K$ does not exceed the estimated validity horizon $\delta_t$; otherwise, it applies the fallback. A live deployment conservatively discretizes end-to-end elapsed time as $K_{\mathrm{live}}=\lceil\Delta/T_{\mathrm{ctrl}}\rceil$. Used only offline, the deterministic reference checker evaluates the same candidate at $x_t$ and $x_{t+K}$; disagreement contributes to the all-candidate verdict-change rate.}
  \label{fig:toctou}
\end{figure*}

We frame this gap as an Execute-stage TOCTOU-style hazard~\cite{bishop1996toctou,lilienthal2025toctou} within the MAPE-K (Monitor--Analyze--Plan--Execute over a shared Knowledge base) loop~\cite{kephart2003vision,ibm2005autonomic} (Fig.~\ref{fig:toctou}). Ordinary closed-loop evolution can invalidate an already-issued approval even without an adversary. Approval latency sets the duration of the exposure window, while plant and predicate dynamics determine whether validity is lost. The central question is not only whether the guardrail returns a correct verdict for context $x_t=(o_t,h_t)$ and action $a_t$ at check time, but whether that verdict remains valid for $(x_{t+K},a_t)$ at actuation. We call validity at use time \emph{verdict freshness} and its loss \emph{verdict staleness}. Its validity therefore depends on when it is used.

The hazard extends beyond LLMs to any delayed Execute-stage approver whose verdict is grounded in an evolving plant state (Sec.~\ref{sec:discussion}). LLM guardrails motivate our study because they insert a model-serving step between planning and actuation, making approval latency part of the control-loop timing. We evaluate the hazard using fixed-action replay over fixed-seed trajectories from five reproducible SAS environments spanning IoT network adaptation, simulator tuning for edge--cloud resource management, architectural self-healing, adaptive object detection, and backend job dispatch: \deltaiot{}~\cite{iftikhar2017deltaiot}, \simtune{}~\cite{simtune2022}, \mrubis{}~\cite{vogel2018mrubis}, \switchenv{}~\cite{marda2024switch}, and \simdex{}~\cite{simdex2022} (Secs.~\ref{sec:design} and~\ref{sec:authenticity}).

We make three contributions. \emph{(i)~Concept:} we identify and formalize an Execute-stage TOCTOU-style hazard in SASs: a semantic approval grounded in an observed plant state can lose validity before actuation under endogenous closed-loop evolution. \emph{(ii)~Measurement:} we hold each proposed action fixed and re-evaluate its admissibility against later recorded contexts, distinguishing all-candidate verdict change, oracle-labeled approval expiry on recorded closed-loop trajectories, and judge-conditioned use-time invalidity. All-candidate verdict change is nonzero in all five environments, and a descriptive audit of four LLM judges finds nonzero judge-conditioned use-time invalidity in every audited approval stream. \emph{(iii)~Mitigation:} we formulate a freshness contract and instantiate it with \FBS{}, a lightweight post-verdict gate that estimates each approval's validity horizon from its safe-side margin and recent feature volatility without an explicit plant-dynamics model. At the common replay shift $K{=}8$, under fixed settings documented in the artifact, \FBS{} reduces oracle-labeled approval-expiry rates in all five environments. In the audited \deltaiot{} cell, it reaches $0\%$ expiry while matching the \refopen{} baseline's reported mean reward.

\section{Related Work: What Becomes Stale?}\label{sec:related}

An Execute-stage verifier must answer two questions: is the proposed action acceptable when checked, and does that approval remain valid when applied? Related work usually addresses check-time acceptability, assigns freshness to a different runtime object, or provides a complementary intervention mechanism.

Runtime LLM guardrails constrain conversational behavior or classify prompts and responses at inspection time~\cite{rebedea2023nemo,inan2023llamaguard}. The GAP benchmark identifies a distinct execution gap: text-level safety need not transfer to the resulting tool call~\cite{cartagena2026gap}. In SAS research, a recent roadmap maps potential LLM roles across MAPE-K, while iLLM-TSC reviews and may revise an action proposed by an RL policy~\cite{li2024roadmap,pang2024illmtsc}. These works address check-time assessment; when such checks authorize execution, they do not estimate how long the resulting approval remains valid under subsequent plant evolution.

Freshness and delay work attaches time to information or adaptation tactics rather than approvals. Age of Information characterizes the age of received status information~\cite{yates2021aoi}, while tactic-volatility work addresses variation in tactic latency or cost~\cite{palmerino2020tactic}. TOCTOU provides the closest structural analogue: classical work examines mutable state between check and use, and recent LLM-agent and browser-agent work revalidates external state before dispatch~\cite{bishop1996toctou,lilienthal2025toctou,jiang2026atomicity}. These lines share the check--use separation, but the revalidated object is information, external state, or authority evidence rather than an Execute-stage semantic approval whose validity changes under endogenous closed-loop plant evolution.

Runtime assurance provides the closest mitigation architecture. Simplex and model-predictive shielding switch to a backup or filter candidate actions based on current or predicted safety~\cite{mehmood2022blackboxsimplex,li2020mpshielding}. \FBS{} follows the same broad intervention pattern but uses a different trigger: it compares an approval's age with a validity horizon estimated from the signed safe-side margin and recent feature volatility, without an explicit plant-dynamics model. We are not aware of prior work that estimates such a plant-dependent horizon for an already-issued Execute-stage semantic approval under endogenous closed-loop evolution. These approaches are complementary reference points rather than direct baselines; comparing them under matched dynamics-model, verifier-call, and fallback assumptions remains future work.

\section{Methodology}\label{sec:design}

Our replay asks one question: if the deterministic reference checker deems candidate action $a_t$, grounded in context $x_t=(o_t,h_t)$, admissible at check time, does the same candidate remain admissible at replay use time $t+K$? To answer it, we hold $a_t$ fixed, take the later recorded context $x_{t+K}=(o_{t+K},h_{t+K})$ from the same episode, and compare the two verdicts of the same deterministic reference checker (the \emph{oracle}). The shifted history is reconstructed from the recorded prefix ending at $t+K$ and contains no later information. This is a fixed-action relabeling audit: it evaluates the candidate against later recorded contexts rather than reconstructing the counterfactual trajectory that delaying, rejecting, or replacing $a_t$ would induce (Sec.~\ref{sec:limitations}).

We use three related quantities with different conditioning sets. The \emph{all-candidate verdict-change rate} averages reference-label changes before conditioning on a method's pass set. The directional \emph{oracle-labeled approval-expiry rate} is computed, for each method, among candidates that method passes and that are reference-admissible at check time in the corresponding recorded experiment. The descriptive LLM audit separately reports \emph{judge-conditioned use-time invalidity} within each judge's own approval set; this quantity may include check-time judge error as well as temporal expiry.

\subsection{System, Threat Model, and Clock Model}

At step $t$, the monitor emits observation $o_t$, the controller proposes action $a_t$, and an Execute-stage LLM guardrail returns verdict $J_t\in\{0,1\}$ on $(x_t,a_t)$, where $x_t=(o_t,h_t)$ includes a short history and $J_t{=}1$ denotes approval. In replay, the same candidate is re-evaluated against $x_{t+K}$. An approval's grounding age is measured from acquisition of $o_t$, not from verdict issuance, and therefore includes all latency between observation and actuation. The threat model excludes prompt injection and deliberate adversarial state manipulation; the hazard is ordinary closed-loop plant evolution during that interval.

We distinguish wall-clock latency $\Delta$ from replay shift $K$ (Fig.~\ref{fig:toctou}). Because the simulators are step-driven, $K$ counts each simulator's own control steps and is the experimental knob. Let $\Delta$ denote the end-to-end elapsed time from acquisition of $o_t$ to actuation and let $T_{\mathrm{ctrl}}$ denote the control period. Replay represents age through integer observation shifts, so we conservatively discretize a latency realization as
\[
K_{\mathrm{live}}=\left\lceil\frac{\Delta}{T_{\mathrm{ctrl}}}\right\rceil .
\]
Thus, any positive remainder beyond an integer number of control periods is assigned to the next age step. For example, a $2$\,s latency and a $0.25$\,s control period yield $K_{\mathrm{live}}{=}8$. Variable serving latency induces a distribution over $K_{\mathrm{live}}$, not a single fixed age.

We sweep $K\in\{1,2,3,5,8,10\}$ and use a common replay shift of eight simulator steps for the headline cross-environment comparison (Table~\ref{tab:ablation} and Fig.~\ref{fig:fbs-impact-all}). This does not imply a common physical duration: step semantics and control periods differ across environments. For each $K$, an index is eligible only when $t+K$ remains within the same episode. The sparse grid covers short, intermediate, and longer observation shifts while keeping the replay campaign tractable. Because the eligible set can shrink with $K$, cross-shift curves are descriptive and may reflect both age and episode-boundary cohort changes (Sec.~\ref{sec:limitations}).

\subsection{Reproducible Environments}

Table~\ref{tab:envs} summarizes the five environments. \deltaiot{} is a multi-hop IoT network whose adaptations adjust per-link communication settings under interference. \simtune{} tunes simulator parameters for edge--cloud resource-management workloads. \mrubis{} is a component-based marketplace exemplar for architectural self-healing. \switchenv{} switches among object-detection model variants as load changes. \simdex{} dispatches jobs across backend workers. The lineup is not intended to make any individual benchmark representative; it applies one replay protocol across systems with different domains, plant dynamics, and operational predicates.

\begin{table}[t]
  \centering
  \scriptsize
  \setlength{\tabcolsep}{1.5pt}
  \caption{The five reproducible SAS environments. ``Admissible'' gives the direction of the selected scalar predicate relative to threshold $\theta$; these predicates provide operational experimental labels rather than complete safety specifications. \switchenv{} uses a per-instance calibrated threshold (``calib.''). Thresholds are held constant across compared methods and seeds.}
  \label{tab:envs}
  \begin{tabular}{P{1.1cm}P{1.25cm}P{1.7cm}P{1.2cm}P{0.85cm}}
    \toprule
    Env & Domain & Gate feature & Admissible & $\theta$ \\
    \midrule
    \deltaiot{}  & IoT network       & \texttt{packet\_loss} & $z\le\theta$ & 0.10 \\
    \simtune{}   & Edge--cloud       & \texttt{rt\_err}      & $z\le\theta$ & 2.0  \\
    \mrubis{}    & Self-healing      & per-step utility      & $z\ge\theta$ & 27.0 \\
    \switchenv{} & Adaptive vision   & \texttt{utility}      & $z\ge\theta$ & calib. \\
    \simdex{}    & Job dispatch      & \texttt{delay\_avg}   & $z\le\theta$ & 5.0  \\
    \bottomrule
  \end{tabular}
\end{table}

\subsection{Replay-Based Freshness Metrics}

Let $g(a_t,x_t;\theta)\in\{0,1\}$ be the binary verdict of the deterministic reference checker, where $g{=}1$ denotes reference admissibility. Let $\mathcal{T}_K$ contain the logged indices for which $t+K$ remains in the same episode, and define $g_t=g(a_t,x_t;\theta)$ and $g_{t+K}=g(a_t,x_{t+K};\theta)$. The all-candidate trajectories use candidates generated by each environment's logged controller, so the resulting rates reflect its controller-induced state, history, and action distribution. For $t\in\mathcal{T}_K$,
\begin{equation}
\mathrm{change}(t,K)=
\mathbf{1}\!\left[g_t\neq g_{t+K}\right].
\label{eq:stale}
\end{equation}
The all-candidate verdict-change rate is the empirical mean of \eqref{eq:stale} over $\mathcal{T}_K$, before conditioning on any evaluated pass set. It includes both expiry ($g_t{=}1,\,g_{t+K}{=}0$) and recovery ($g_t{=}0,\,g_{t+K}{=}1$). The term \emph{all-candidate} describes the denominator; the rate remains conditional on the logged distribution.

For a method $M$, let $S_t^M\in\{0,1\}$ indicate that $M$ passes candidate $a_t$ in the corresponding recorded experiment. Its directional oracle-labeled approval-expiry rate is
\begin{equation}
R_{\mathrm{exp}}(M,K)=\Pr\!\left[g_{t+K}{=}0\mid S_t^M{=}1,\,g_t{=}1\right].
\label{eq:expiry}
\end{equation}
This quantity asks: among candidates passed by $M$ that were reference-admissible at check time, what fraction are inadmissible at replay use time? Recovery is a missed opportunity rather than an unsafe stale approval. Paired seeds align environment randomness across compared methods; the metric does not assume that their candidate streams remain identical after fallback interventions.

The descriptive LLM audit uses a different denominator. For judge verdict $J_t$, it reports
\begin{equation}
R_{\mathrm{judge}}(J,K)=\Pr\!\left[g_{t+K}{=}0\mid J_t{=}1\right].
\label{eq:judge}
\end{equation}
Because \eqref{eq:judge} does not additionally condition on $g_t{=}1$, it may include both candidates already inadmissible at check time and candidates whose admissibility expires later. It therefore measures judge-conditioned use-time invalidity, not temporal expiry alone.

The LLM judge, reference checker, and \FBS{} play different roles. The judge supplies the check-time semantic verdict. The deterministic reference checker is an offline measurement device, never deployed in the loop, that labels admissibility at check and replay use time. \FBS{} decides whether an upstream approval remains within an estimated temporal horizon; it does not re-judge the candidate's semantics. Calling the checker an oracle is shorthand for its measurement role, not a claim that it captures every safety-relevant property. The conceptual contribution concerns the lifetime of a semantic approval, whereas the main timing experiment uses deterministic operational labels and \FBS{} uses a scalar proxy; the separate LLM audit does not constitute end-to-end Judge$+$\FBS{} evaluation.

\subsection{Freshness-Bounded Shield}

Each adapter exposes a scalar state feature $z_t=f(o_t)$ used by the freshness gate. The offline reference checker may additionally use $a_t$ and $h_t$; accordingly, \FBS{} is a lightweight proxy for the selected scalar predicate rather than a certificate for the full checker. Define the signed safe-side margin as $m_t=\theta-z_t$ when admissibility requires $z\le\theta$, and $m_t=z_t-\theta$ when it requires $z\ge\theta$. Thus, $m_t\ge0$ on the admissible side and $m_t>0$ strictly inside the boundary. The estimated horizon is
\begin{equation}
\begin{aligned}
\delta_t &= \min\!\left(\delta_{\max},\ \frac{\max(m_t,0)}{\max(\mathrm{vol}_t,\varepsilon)}\right),\\
\mathrm{vol}_t &= \alpha\,|z_t-z_{t-1}|+(1-\alpha)\,\mathrm{vol}_{t-1},
\end{aligned}
\label{eq:fbs}
\end{equation}
with $\alpha{=}0.1$ fixed across all five environments. At this value, the contribution of an observed change to the EMA decays by half after $\ln(1/2)/\ln(0.9)\approx6.6$ simulator steps. The margin measures distance to the scalar boundary, and $\mathrm{vol}_t$ is an exponential moving average of recent absolute feature changes. Their ratio is used as a heuristic time-to-boundary estimate. The floor $\varepsilon>0$ prevents a momentarily static feature from receiving an unbounded horizon, while $\delta_{\max}$ caps the horizon. The artifact fixes $\varepsilon$, $\delta_{\max}$, initialization, and update order.

Let $A_t\in\{0,1\}$ denote the upstream approval. In a deployed Judge$+$\FBS{} composition, $A_t$ is the judge verdict; in the oracle-labeled timing experiment, $A_t=g_t$. The candidate passes if and only if $A_t{=}1$, $m_t>0$, and $K\le\lfloor\delta_t\rfloor$. Otherwise, the environment-specific fallback is applied.

The rule is motivated by a bounded-drift argument for the selected scalar predicate. If $|z_{s+1}-z_s|\le b$ at every step, then the feature moves by at most $Kb$ over $K$ steps, and $Kb\le m_t$ is sufficient to avoid crossing an inclusive threshold. \FBS{} substitutes the smoothed recent change $\mathrm{vol}_t$ for the unknown worst-case bound $b$. Because $\mathrm{vol}_t$ is not an upper bound, the resulting horizon is a heuristic rather than a safety certificate: an abrupt change can outpace the estimate and invalidate a passed approval. This is a possible failure mode of the rule, not evidence that it caused any particular residual observed below. The signed margin also prevents a large distance on the inadmissible side from producing a long lifetime; if $m_t\le0$, \FBS{} invokes the fallback.

\subsection{Methods Compared}

Each compared variant is evaluated under paired seeds in its corresponding recorded closed-loop experiment. The \refopen{} baseline applies every candidate with positive upstream approval and performs no freshness check. In the oracle-labeled timing experiment, $A_t=g_t$, so it passes every candidate deemed admissible by the reference checker at check time. \FBS{} (\efbs{}) uses the same upstream approval source in that experiment and additionally rejects candidates whose estimated horizon has expired. Thus, the reported \eopen{}--\FBS{} comparison isolates freshness gating; it is not an end-to-end evaluation of an LLM-generated approval stream. \emarg{} retains only boundary-distance information, while \evol{} retains only recent-volatility information. \edeadline{} rejects every approval beyond a fixed age regardless of context and appears only in the \deltaiot{} safety--utility audit. Exact ablation rules, deadline, parameter values, fallback definitions, and pass frequencies are recorded in the artifact.

\subsection{Reproducibility and Traceability}\label{sec:authenticity}

All five environments pass a two-run reproducibility test under fixed seeds. Reproduction is bit-exact where supported and tolerance-bounded for JVM-backed environments whose upstream simulators reproduce floating-point outputs within a fixed numerical tolerance rather than bit for bit. Each adapter imports directly from the corresponding upstream simulator, and a SHA-256 manifest pins the artifacts and generated traces. Closed-loop comparisons use $n{=}30$ paired seeds per environment and method ($n{=}60$ for \simdex{}). Decision-level outcomes are aggregated into one rate per seed, and methods are compared using paired same-seed Wilcoxon signed-rank tests~\cite{wilcoxon1945}. The artifact records the SciPy version, the exact \texttt{alternative}, \texttt{zero\_method}, \texttt{method}, and continuity-correction settings, the treatment of tied and zero differences, the handling of seeds with no passed candidates, the prompts, decoding settings, model identifiers, and per-seed outputs.

\section{Results}\label{sec:results}

We address three research questions. \textbf{RQ1:} How does all-candidate verdict change vary across environments and replay shifts? \textbf{RQ2:} At $K{=}8$, does \FBS{} reduce oracle-labeled approval expiry on recorded closed-loop trajectories, and, in the audited \deltaiot{} cell, what do the ablation and safety--utility audit reveal about over-blocking? \textbf{RQ3:} What judge-conditioned use-time invalidity remains in concrete LLM approval streams?

All three analyses re-label admissibility at replay use time, but they condition on different sets. The curves in Fig.~\ref{fig:fbs-impact-all} report all-candidate verdict change over eligible logged candidates. Table~\ref{tab:ablation} and the $K{=}8$ markers in Fig.~\ref{fig:fbs-impact-all} report method-specific approval expiry among passed candidates that are reference-admissible at check time in each corresponding experiment. Fig.~\ref{fig:llm} instead reports $R_{\mathrm{judge}}(J,K)$ within each judge's own approval set. The direct mitigation comparison is therefore \refopen{} versus \FBS{} within the oracle-labeled timing experiment. It isolates freshness gating and should not be interpreted as an end-to-end comparison of LLM approval streams; numerical gaps across the three metric families are not effect estimates.

\subsection{RQ1: How Does Verdict Change Vary?}

All-candidate verdict change is nonzero in every environment. At the common replay shift $K{=}8$, it ranges from $5.3\%$ in \simdex{} to $48.4\%$ in \deltaiot{}, a roughly ninefold spread (Fig.~\ref{fig:fbs-impact-all}). Because a simulator step need not represent the same physical duration across environments, this is a common integer shift rather than a common wall-clock age. Plant dynamics, predicate structure, and the logged candidate distribution determine how often the reference verdict changes within each environment's shift.

The curves exhibit four descriptive shapes. \deltaiot{} and \simtune{} are \emph{early-saturating}: their rates are already near $48\%$ at $K{=}1$ and vary little over the sampled ages. This pattern is consistent with frequent boundary crossings or weak temporal persistence, but the curves alone do not identify the cause. \mrubis{} is \emph{age-sensitive}: its rate rises overall from $16.5\%$ at $K{=}1$ to $32.7\%$ at $K{=}8$, despite local non-monotonicity. \switchenv{} is approximately \emph{flat}, while \simdex{} is \emph{low-base} with small local variation. These labels summarize observed shapes rather than establish underlying mechanisms.

\subsection{RQ2: Does \FBS{} Reduce Approval Expiry?}

\begin{figure*}[t]
  \centering
  \includegraphics[width=0.84\textwidth]{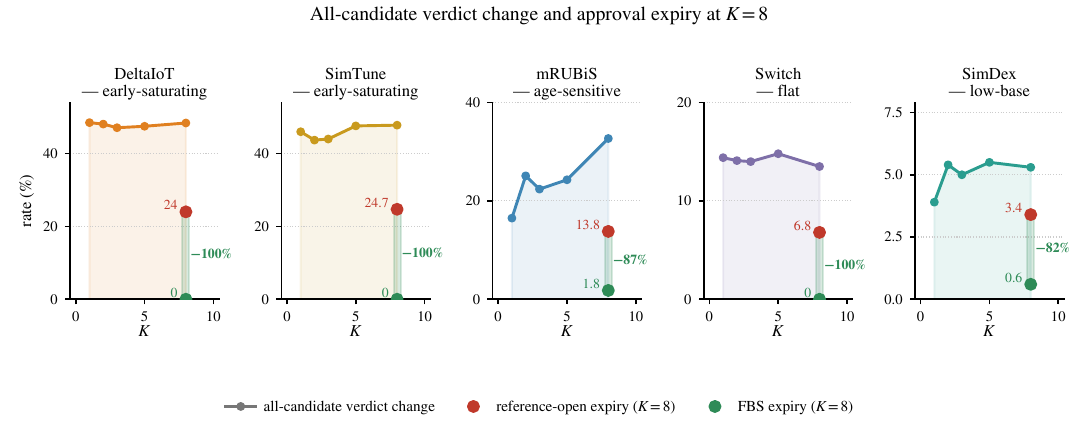}
  \caption{All-candidate verdict change across replay shift $K$, with method-specific oracle-labeled approval expiry at $K{=}8$. Each curve reports the all-candidate verdict-change rate for one environment. The red and green markers report approval-expiry rates for \refopen{} and \FBS{} in their corresponding recorded experiments. Curves and markers represent different events and denominators; curve--marker gaps are not effect estimates. The direct mitigation comparison is between the paired red and green outcomes. $K$ counts each simulator's own steps, not a common wall-clock duration.}
  \label{fig:fbs-impact-all}
\end{figure*}

At $K{=}8$, \FBS{} reports a lower oracle-labeled approval-expiry rate in all five environments, reducing the \refopen{} range of $3.4$--$24.7\%$ to $0$--$1.8\%$ (Fig.~\ref{fig:fbs-impact-all} and Table~\ref{tab:ablation}). The observed \FBS{}-to-\eopen{} rate ratios range from $0$ to $0.18$, equivalent to relative reductions of $82$--$100\%$. All five paired per-seed comparisons yield raw $p{<}0.001$ under the configured Wilcoxon signed-rank tests. Because each rate conditions on the candidates passed by the corresponding method, a lower expiry rate alone does not determine whether the reduction is selective or results from more aggressive rejection.

\subsection{Ablation: Freshness Signals and Over-Blocking}

The single-signal variants isolate the two inputs to \eqref{eq:fbs}, but approval expiry alone cannot identify over-blocking because each variant induces a different pass set. \emarg{} uses only distance to the admissibility boundary and reaches $0\%$ approval expiry in every environment; the expiry table alone cannot determine whether this reflects selective gating or conservative rejection. \evol{} uses only recent feature movement and remains close to \refopen{} in \deltaiot{}, \simtune{}, and \switchenv{}; for example, their rates are $23.9\%$ and $24.0\%$ on \deltaiot{}. On \simdex{}, \evol{} matches the full-\FBS{} rate of $0.6\%$. These patterns motivate combining boundary distance and recent motion, but do not by themselves establish a safety--utility benefit.

The \deltaiot{} safety--utility audit at $K{=}8$ provides complementary evidence in one cell (Table~\ref{tab:safety-utility}). \emarg{} reaches $0\%$ approval expiry but reduces reported mean episode reward from the \refopen{} baseline's $-960$ to $-1050$, while \edeadline{} reduces it to $-1200$. \FBS{} reaches $0\%$ approval expiry while matching the reported reference-open mean reward of $-960$. Thus, in this audited cell, \FBS{} reduces expiry without the mean-reward loss observed for the two more conservative alternatives. This does not establish the same trade-off in other environments, at other replay shifts, or under other fallback designs.

\begin{table}[t]
  \centering
  \scriptsize
  \caption{Oracle-labeled approval-expiry rate (\%) at $K{=}8$ on recorded closed-loop trajectories. Each value conditions on candidates passed by the corresponding variant and reference-admissible at check time. Lower values alone do not establish a better safety--utility trade-off.}
  \label{tab:ablation}
  \setlength{\tabcolsep}{2.8pt}
  \begin{tabular}{P{1.65cm}ccccc}
    \toprule
    Variant & \deltaiot{} & \simtune{} & \mrubis{} & \switchenv{} & \simdex{} \\
    \midrule
    Ref.-open (\eopen{}) & 24.0 & 24.7 & 13.8 & 6.8 & 3.4 \\
    \emarg{} & 0.0 & 0.0 & 0.0 & 0.0 & 0.0 \\
    \evol{} & 23.9 & 24.0 & 8.9 & 6.7 & 0.6 \\
    \FBS{} & 0.0 & 0.0 & 1.8 & 0.0 & 0.6 \\
    \bottomrule
  \end{tabular}
\end{table}

\begin{table}[t]
  \centering
  \scriptsize
  \caption{Safety--utility audit on \deltaiot{} at $K{=}8$. Approval expiry conditions on candidates passed by the method and reference-admissible at check time; reward is the reported mean episode reward over paired seeds. }
  \label{tab:safety-utility}
  \setlength{\tabcolsep}{3.2pt}
  \ifshowfallback
\else
  \begin{tabular}{lcc}
    \toprule
    Method & Expiry (\%) & Mean reward \\
    \midrule
    Ref.-open (\eopen{}) & 24.0 & $-960$ \\
    \FBS{} & 0.0 & $-960$ \\
    \emarg{} & 0.0 & $-1050$ \\
    \evol{} & 23.9 & $-965$ \\
    \edeadline{} & 0.0 & $-1200$ \\
    \bottomrule
  \end{tabular}
  \fi
\end{table}

\subsection{RQ3: Use-Time Invalidity in LLM Approval Streams}

Every audited judge stream has a nonzero observed use-time invalidity rate. We replayed approval streams from four LLM judge backends---\texttt{qwen2.5:0.5b}, \texttt{llama3.2:1b}, Claude Haiku, and \texttt{gpt-4o-mini}---under the same observation-shifting protocol. Each judge receives $(x_t,a_t)$ and the environment's operational admissibility predicate and returns an approve/reject verdict.

\begin{figure}[!tb]
  \centering
  \includegraphics[width=0.86\linewidth]{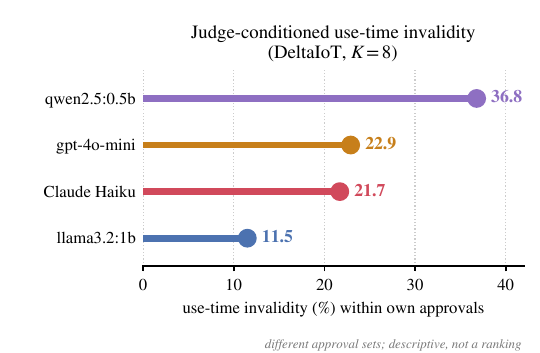}
  \caption{Judge-conditioned use-time invalidity in four LLM approval streams on \deltaiot{} at $K{=}8$. Each marker is computed within that judge's own approval set and may include both check-time judge error and temporal expiry. The observed rates range from $11.5\%$ to $36.8\%$; because the approval sets differ, marker heights characterize separate streams rather than rank judge quality.}
  \label{fig:llm}
\end{figure}

At $K{=}8$ in the early-saturating \deltaiot{} setting, use-time invalidity within each backend's own approval set ranges from $11.5\%$ to $36.8\%$ (Fig.~\ref{fig:llm}). Because each judge induces a different approval set and the audit does not additionally condition on reference admissibility at check time, these rates may combine check-time judge error with temporal expiry. They characterize exposure within each stream; they do not isolate temporal expiry, rank judge quality, or provide a direct comparison with \FBS{}, which is evaluated separately. The full backend-by-environment matrix is provided in the artifact. End-to-end Judge$+$\FBS{} evaluation on the same judge-generated stream remains future work.

\section{Discussion}\label{sec:discussion}

For deployment, check-time judge accuracy and median verifier latency are not enough. The latency distribution determines the grounding-age distribution, while signed margin and recent feature volatility help characterize exposure to plant drift~\cite{dean2013tail,yates2021aoi}. Runtime monitoring should track the distributions of $K_{\mathrm{live}}$, $m_t$, and $\mathrm{vol}_t$, together with fallback frequency and task utility. Periodic offline replay audits should report oracle-labeled approval expiry separately. Judge-conditioned use-time invalidity is a different quantity because it conditions on each judge's own approval set and may include check-time error.

These requirements define a \emph{freshness contract}: an approval records its checked context and action, grounding time, expiry or revalidation rule, and a fallback justified for the deployment. \FBS{} implements one version of this contract using signed margin and recent feature volatility, without an explicit dynamics model; other deployments may use different mechanisms.

Several complementary mechanisms can reduce the hazard. Lower verifier latency narrows but does not eliminate the check--use window; a dynamics model can support prediction of future safety or recoverability~\cite{li2020mpshielding}, while revalidation immediately before dispatch can check a more current state~\cite{jiang2026atomicity}. \FBS{} instead expires an earlier approval without another verifier call, but its horizon is heuristic rather than a safety certificate. Its value depends on the fallback: a no-op or held action is not inherently safe. Because approval-expiry rates condition on each method's pass set, they should be reported with fallback frequency and task utility~\cite{mehmood2022blackboxsimplex}. The \deltaiot{} audit shows one favorable safety--utility point, not a general guarantee.

Once $z_t$, $\theta$, and age are available, \FBS{} adds $O(1)$ scalar work and no additional model call; this excludes deployment-specific feature extraction, timestamping, and fallback execution. The freshness requirement is not LLM-specific, but applying \FBS{} elsewhere requires a meaningful scalar margin, a usable online volatility estimate, trustworthy age measurement, and a justified fallback. End-to-end evaluation of Judge$+$\FBS{} on the same approval stream remains future work.

\section{Limitations}\label{sec:limitations}

This study evaluates all-candidate verdict change and oracle-labeled approval expiry in five step-driven simulators at a sparse set of replay shifts $K$, and evaluates \FBS{} under fixed settings ($\alpha{=}0.1$; the remaining parameters are fixed in the artifact). Equal $K$ values denote equal simulator-step shifts, not equal physical durations. The findings apply to the evaluated environments, scalar predicates, logged candidate distributions, replay grid, and configuration, not to a particular deployment. A deployment-specific estimate requires measured end-to-end latency traces, their discretization into $K_{\mathrm{live}}$, and evaluation on the target plant and controller-induced action distribution. Sensitivity to \FBS{} parameters and unsampled ages remains unmeasured. Because the eligible replay cohort shrinks near episode boundaries as $K$ increases, cross-age curves may also reflect cohort changes.

Replay is a fixed-action relabeling audit, not an intervention-consistent causal simulation: it evaluates $a_t$ on later recorded contexts rather than reconstructing the trajectory induced by delaying, rejecting, or replacing it. The reference checker supplies deterministic labels for selected operational predicates but is not a complete safety oracle. \FBS{} uses a scalar state feature as a proxy and therefore does not certify every dependency of the full checker. Utility and fallback trade-offs are tested only on \deltaiot{} at $K{=}8$. Judge-conditioned rates may mix check-time error with temporal expiry; check-time LLM accuracy and same-stream Judge$+$\FBS{} composition are not evaluated. Finally, \FBS{} replaces an unknown worst-case drift bound with a smoothed recent-change estimate and depends on trustworthy age measurement and a justified fallback. It is a proof-of-concept heuristic, not a certified safety guarantee.

\section{Conclusion}\label{sec:conclusion}

An Execute-stage approval is not a timeless Boolean. It can lose validity as the plant evolves away from the context on which it was based. Across five reproducible, fixed-seed SAS environments, the all-candidate verdict-change rate is nonzero and spans $5.3$--$48.4\%$ at the common replay shift $K{=}8$. This roughly ninefold spread shows that age alone does not determine the observed rate; plant dynamics, predicate structure, and the logged candidate distribution also matter.

Under fixed settings documented in the artifact, \FBS{} reduces oracle-labeled approval-expiry rates on recorded closed-loop trajectories in all five environments. In the audited \deltaiot{} cell at $K{=}8$, it reaches $0\%$ approval expiry while matching the \refopen{} baseline's reported mean reward. The separate LLM audit finds nonzero judge-conditioned use-time invalidity in all four approval streams, although it does not isolate temporal expiry from check-time judge error.

\FBS{} remains a proof-of-concept heuristic. Deployment-specific evaluation and calibration, together with end-to-end Judge$+$\FBS{} evaluation on the same approval stream, remain future work. The broader contribution is the \emph{freshness contract}: semantic approval at check time must be paired with an explicit validity-at-use rule and a justified fallback on expiry. An Execute-stage assurance mechanism that establishes only check-time correctness answers only half of the safety question.

\bibliographystyle{IEEEtran}
\bibliography{references}
\end{document}